# A Study on AI-FML Robotic Agent for Student Learning Behavior Ontology Construction

Chang-Shing Lee, Mei-Hui Wang, Wen-Kai Kuan
Zong-Han Ciou, Yi-Lin Tsai, Wei-Shan Chang
Dept. of Computer Science and Information Engineering
National University of Tainan
Tainan, Taiwan
leecs@mail.nutn.edu.tw

Lian-Chao Li, Naoyuki Kubota
Dept. of Mechanical Systems Engineering
Tokyo Metropolitan University
Tokyo, Japan
kubota@tmu.ac.jp

Tzong-Xiang Huang, Eri Sato-Shimokawara
Toru Yamaguchi
Dept. of System Design
Tokyo Metropolitan University
Tokyo, Japan
yamachan@tmu.ac.jp

***Abstract*—In this paper, we propose an AI-FML robotic agent for student learning behavior ontology construction which can be applied in English speaking and listening domain. The AI-FML robotic agent with the ontology contains the perception intelligence, computational intelligence, and cognition intelligence for analyzing student learning behavior. In addition, there are three intelligent agents, including a perception agent, a computational agent, and a cognition agent in the AI-FML robotic agent. We deploy the perception agent and the cognition agent on the robot Kebbi Air. Moreover, the computational agent with the Deep Neural Network (DNN) model is performed in the cloud and can communicate with the perception agent and cognition agent via the Internet. The proposed AI-FML robotic agent is applied in Taiwan and tested in Japan. The experimental results show that the agents can be utilized in the human and machine co-learning model for the future education.**

## I. Introduction

What is the relationship between Artificial Intelligence (AI) and Computational Intelligence (CI)? According to IEEE Computational Intelligence Society (CIS), CI is the theory, design, application and development of biologically and linguistically motivated computational paradigms [1]. It plays an important role in developing successful intelligent systems, including games, multilayer perceptron, and cognitive developmental systems, for example, Deep Learning (DL) or Deep Convolutional Neural Networks (DCNN). DL or DCNN is one of the core methods for AI systems [1]. In addition, using the human language as a source of inspiration, fuzzy logic systems (FLS) can model linguistic imprecision and solve uncertain problems [1]. FLS enables us to perform approximate reasoning based on humanized thinking, and it includes fuzzy sets and fuzzy logic, fuzzy clustering and classification, linguistic summarization, fuzzy neural networks, type-2 fuzzy sets and systems, and so on [1].

B. Bouchon-Meunier, President of the IEEE CIS in 2020-2021, wrote her welcome message in the IEEE CIS Website to the readers as follows [2]: "…One century ago, in 1920, the word "robot" appeared, … 1920 is also the year when the writer Isaac Asimov was born and everyone knows his *Three Laws of Robotics*, which he first published in 1942 in the novel *Runaround*. This novel was the first in a long series of stories in which he explored relations between humans and artifacts and questioned the applicability of these rules to robots in real-life situations with humans….", and, "…Claude E. Shannon published his paper "*Programming a computer for playing chess*," the first paper proposing a computer program to play chess, a distant ancestor to AlphaGo. The same year, in 1950, Alan Turing published "*Computing Machinery and Intelligence*," in which he proposed to consider the question "Can machines think?" and described the "imitation game", now known as the Turing test, a "big leap for mankind", from my point of view.".

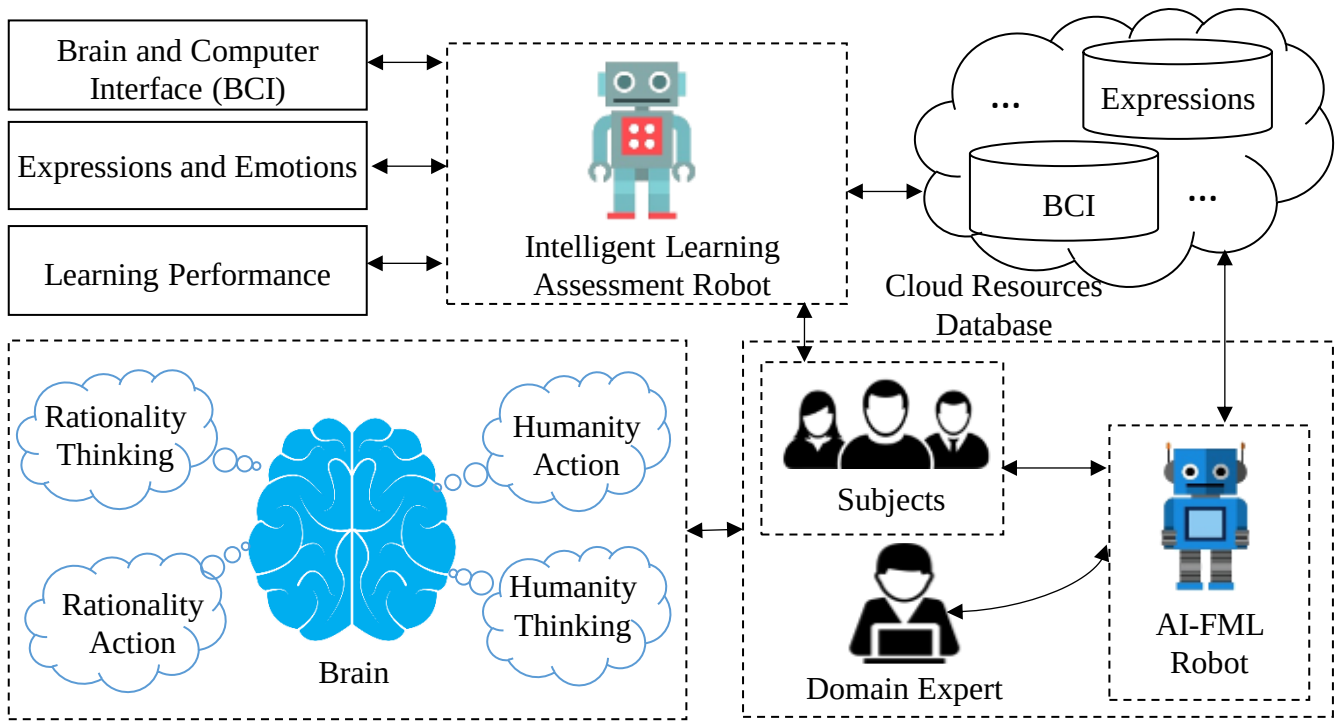

Fig. 1. System structure for student learning performance assessment with BCI mechanism [7].

Fig. 1 shows the system structure for student learning performance assessment with brain computer interface (BCI) mechanism which describes as follows [7]: The subject wearing a BCI device interacts with the intelligent learning assessment robot and AI-FML robot to learn English through listening or speaking [7]. Meanwhile, the data related to his/her expressions, emotions, and learning performance are collected to store in the cloud resources/database. The left side of the brain is better at things like reading, writing, and computations [3]. However, the right brain has a more creative and less organized way of thinking [3]. Hence, the robot can help humans to do things in a more rational thinking, and human can add much creative information and knowledge to the robotic behavior. Therefore, humans can work with the robot in the world such as listening to music or Go applications for human and machine co-learning in future education [7-10]. In this paper, we propose an AI-FML robotic agent for student behavior ontology construction and analyze student learning behavior. Kebbi Air robots with AI-FML robotic agent were deployed in Taiwan and tested in Japan.

The research question in this paper is try to observe student learning behavior in class to construct their learning behavior ontology. Fuzzy markup language (FML) is a human machine language and it is able to establish the communication bridge between humans and machines. Humans interact with the robots to make learning fun and the robots collect the information from humans to make themselves much smarter after machine learning. That's way, we use FML as a communication language between humans and machines. The experimental results show the agents can be utilized in the human and machine co-learning model for the future education.

The remainder of the paper is organized as follows. We first introduce the system structure of AI-FML robotic agent in Section II. We then analyze and construct the student learning behavior extraction mechanism based on the robotic agent in Section III. Section IV presents the exploration and exploitation for learning behavior ontology construction. Experimental results are given in Section V and finally we conclude the paper in Section VI.

## II. System Structure of AI-FML Robotic Agent

### A. *AI-FML Agent with Perception Agent, Computational Agent and Cognition Agent*

In this paper, we classify the intelligence of AI-FML into three parts, including *Perception Intelligence*, *Cognition Intelligence*, and *Computational Intelligence*. Fig. 2 shows the structure of AI-FML robotic agent which is deployed to the robot Kebbi Air developed by NUWA Robotics. The specification of the Kebbi Air is 355mm long × 347mm wide×196mm high and its weight is 4400g. Containing neural networks, fuzzy systems, and evolutionary computation, the computational intelligence can be applied to many real-world applications, for example, to ambient intelligence, artificial life, cultural learning, computer games, and social reasoning [1].

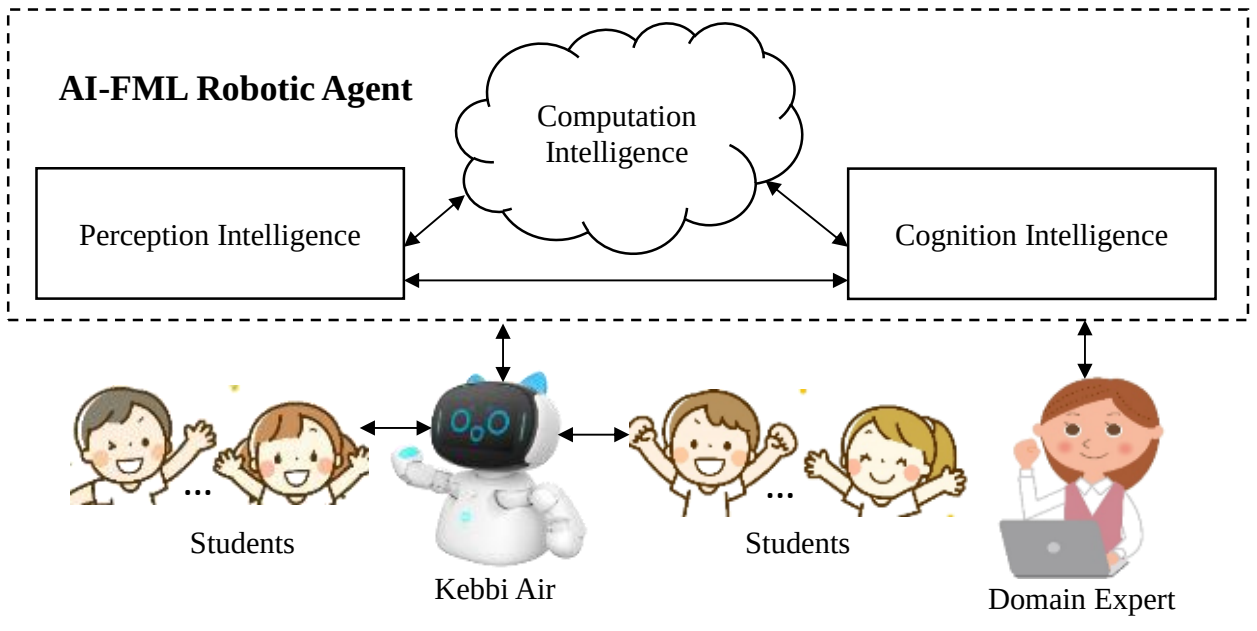

Fig. 2. Structure of AI-FML robotic agent with perception intelligence, cognition intelligence, and computational intelligence.

The AI-FML robotic agent contains the *perception agent*, the *cognition agent*, and the *computational agent*. The perception agent with perception intelligence has the ability to get data, information, or knowledge from the environment based on some devices and perceptron through voice recognition, image processing, or natural language processing. It can help a machine to communicate with a person to deal with real-world applications. The cognition agent with cognition intelligence is able to receive information or knowledge from the environment, perception agent, or the computational agent. Then, it generates various specific responses to the environment based on each human experience and personality. For example, Kebbi Air can infer the student learning performance according to the retrieved learning information in the past.

### B. *Perception Agent and Cognition Agent Structure*

Before learning happens, humans first need to have input. Second, humans transform the information into a model that we use to take action on our surroundings. Finally, learning happens after humans perform the action to new information [4]. Fig. 3 shows the structure of the perception agent and cognition agent in the teaching fields where students learn through the interaction with the various kinds of the robots and human teachers observe their students' learning performance in different learning pathways to human and robot co-learning. Here, the perception agent is responsible for gathering data and information using its sensors, and the cognition agent observes the involved students' behavior of human actions and human thinking.

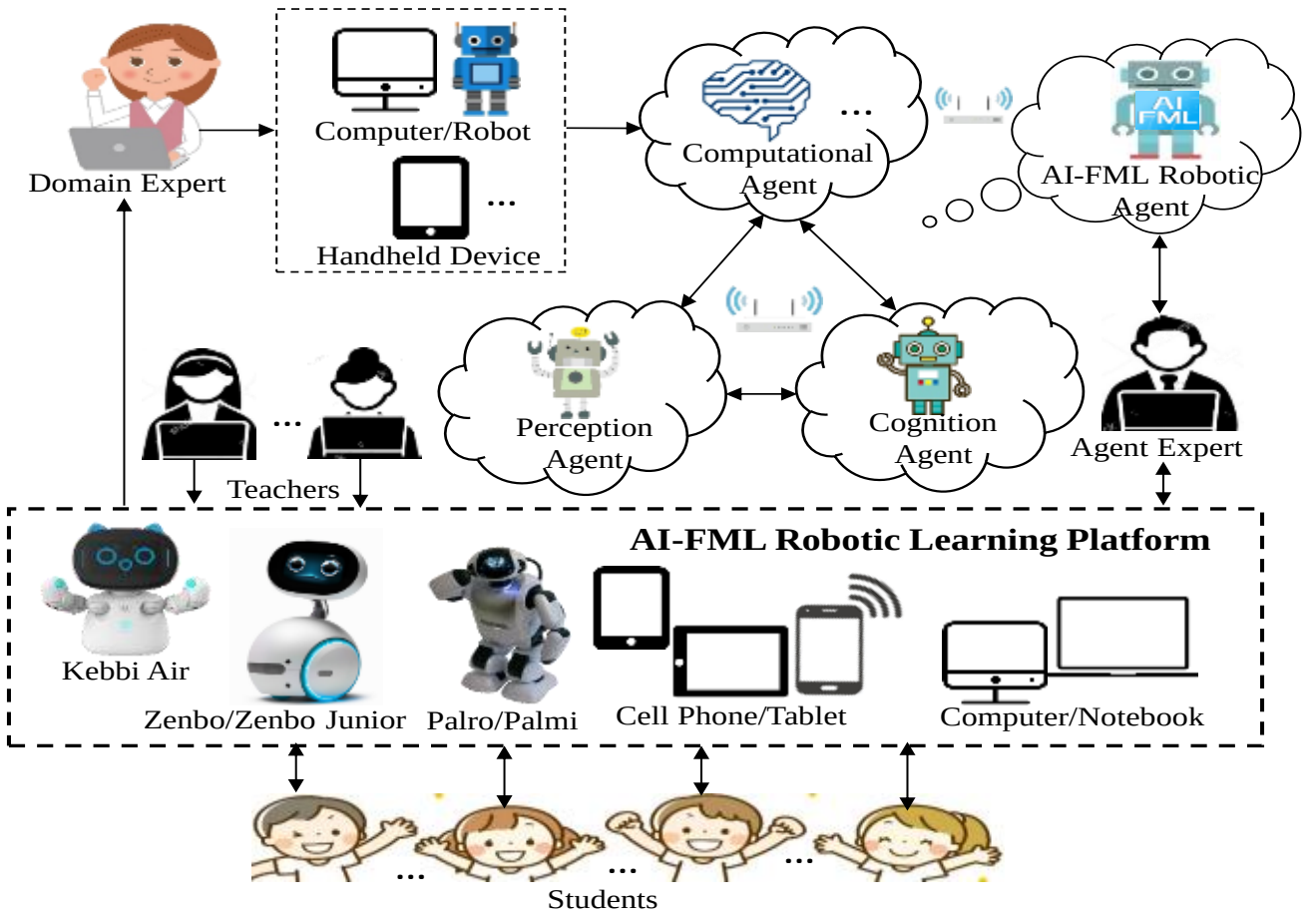

Fig. 3. System structure of the perception agent and cognition agent.

### C. *Computational Agent with Deep Neural Network Structure*

Fig. 4 shows the structure of the computational agent with deep neural network for AI-FML robotic agent.

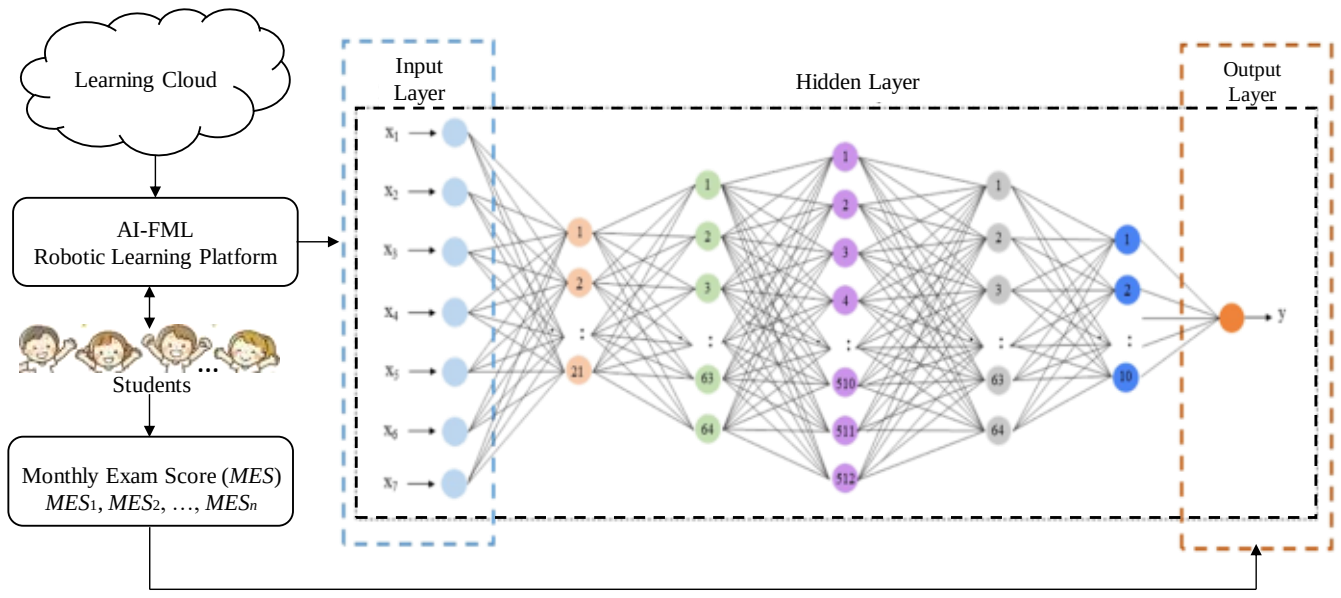

Fig. 4. Structure of the computational agent with deep neural network for AI-FML robotic agent.

The involved students learn the teaching materials on the cloud through the AI-FML robotic learning platform and their learning performance are stored in the cloud. Feature variables, for example, $x_1$, $x_2$, …, and $x_7$, of the deep neural network are extracted from the collected data during the learning period and $y$ is the output of the established deep learning model. The labeled variable could be the involved students' monthly exam score (*MES*), for example, $MES_1$, $MES_2$, .., $MES_n$, in the semester. The learned model is used to predict their learning performance to adjust the learning method of the involved students in the future. The detailed information about the the network architecture of deep neural network for AI-FML robotic agent, including the input layer, hidden layers, and output layer, are given in Section V.A.

## III. AI-FML Robotic Agent Analysis and Construction for Student Learning Behavior Extraction Mechanism

### A. *Use Case Diagram for Learning Behavior Extraction*

Fig. 5 shows a use case diagram for learning behavior extraction. Here, we use an intelligent speaking assistance system, including a student sub-system, a teacher sub-system, and a backend sub-system, as an example. The backend sub-system provides a user interface to allow human teachers to import the involved students' profile, the edited questions, and so on. The involved teachers edit the questions and validate the questions using the developed teacher sub-system. After that, a student makes a challenge to practice speaking to collect his/her challenge result. Additionally, the student sub-system provides the playback function to play the sound of text recognized by AI. Finally, their teachers can view their students' learning performance after they login the system. The collected information contains such as "*do they practice speaking*," "*how many times they practice*," and *"how well their speaking performance*." We extract the involved students' learning behavior according to the collected information.

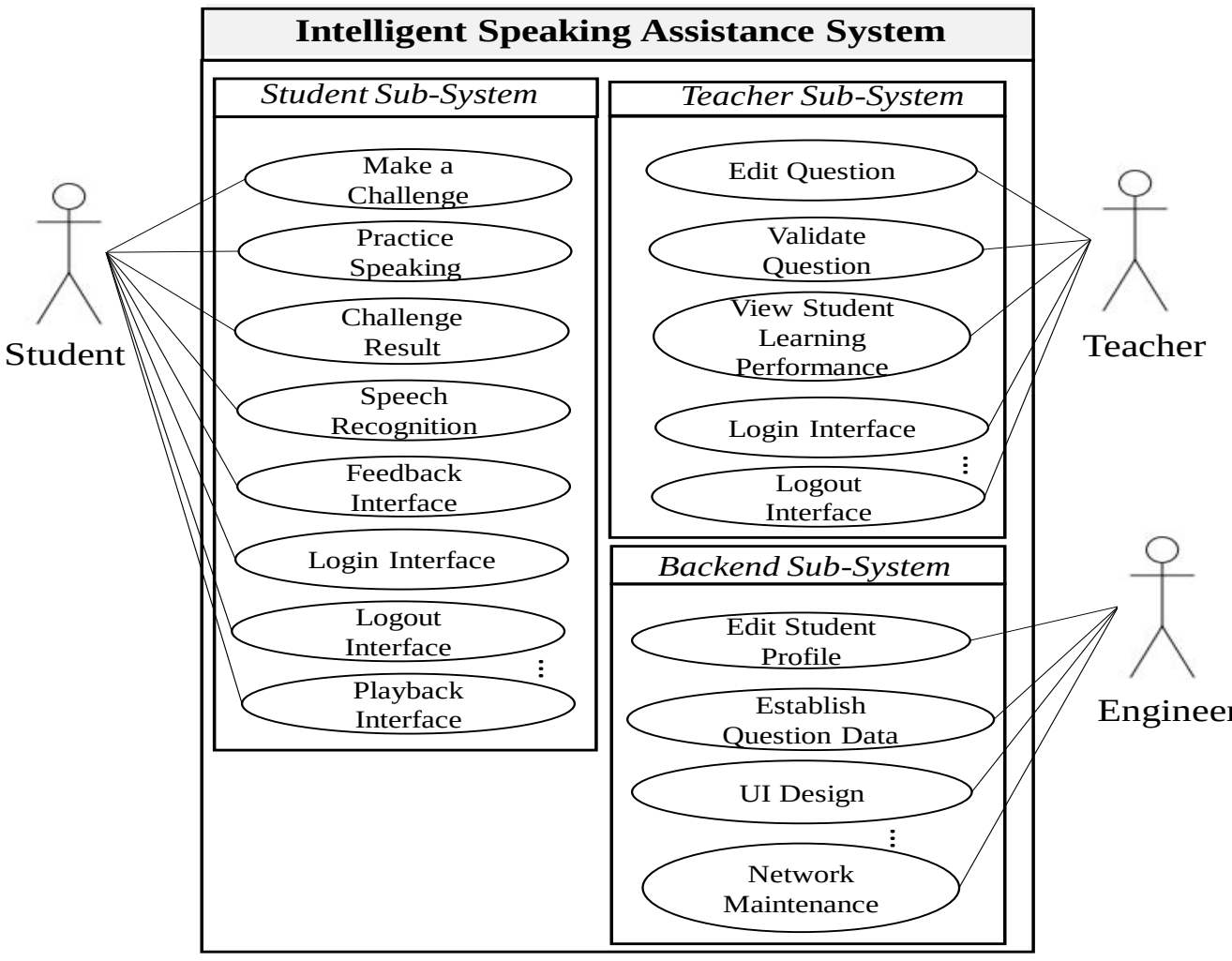

Fig. 5. Use case diagram of the intelligent speaking assistance system.

### B. *Work Breakdown Structure for Perception Agent and Cognition Agent*

Fig. 6 shows the work breakdown structure (WBS) of the intelligent speaking assistance system, including six sub-deliverables work packages (WPs) for developing three sub-systems to extract the involved students' perception data and cognition information during learning. Each WP has the list of tasks or "to-dos" to produce the specific unit of work. For example, requirement analysis (WP1) needs to write down requirement analysis (T1.1) and conduct expected effect analysis (T1.2). In this paper, we develop an intelligent speaking assistance system to deploy it to the teaching fields in Taiwan and tested in Japan. Hence, we should take funding, process, equipment, feedbacks, human resources, and place related to teaching fields into consideration in addition to system design, implementation, and integration related to the student, teacher, and backend sub-systems. Finally, we should have human resources to deploy them to the teach fields to assist in human teachers in class in the initial introducing phase.

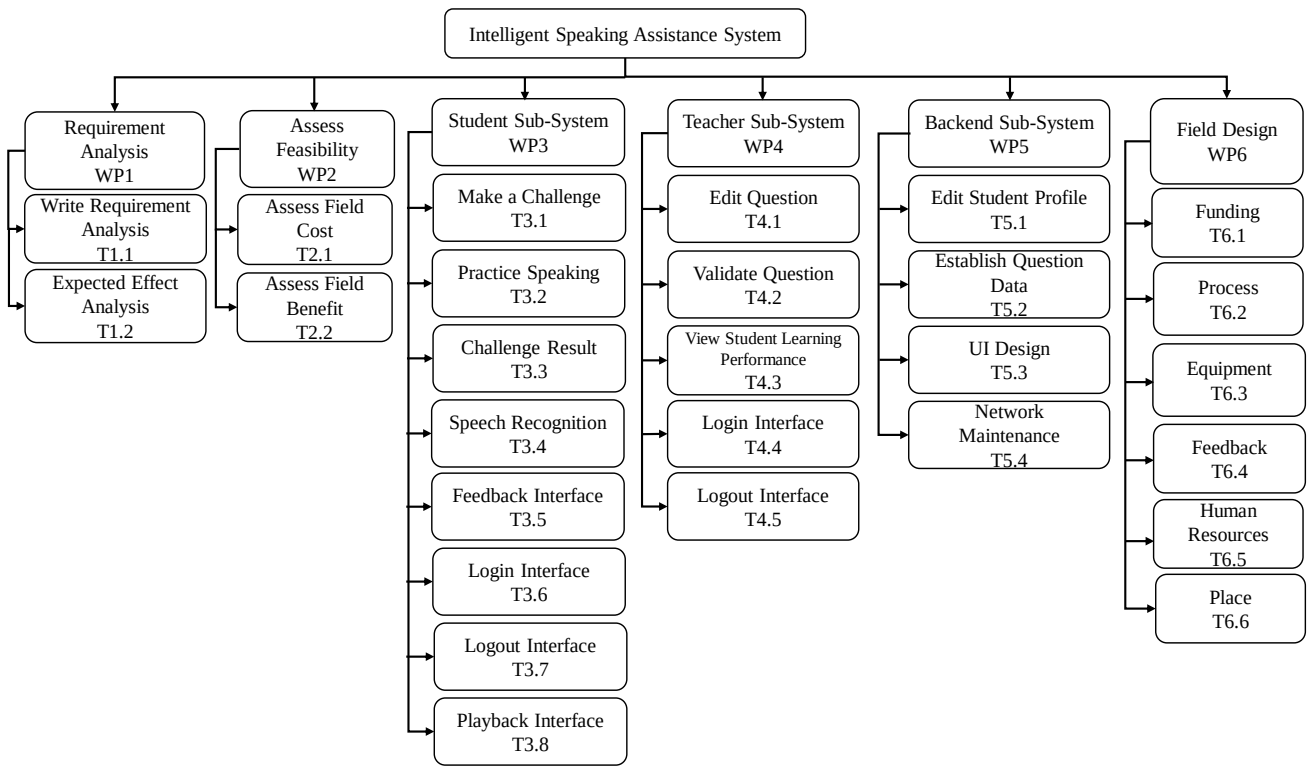

Fig. 6. WBS of the intelligent speaking assistance system.

### C. *User Interface and Functionalities for Learning Behavior Extraction*

We develop the user interface and functionalities for behavior extraction based on the above-mentioned use case diagram and WBS structure.

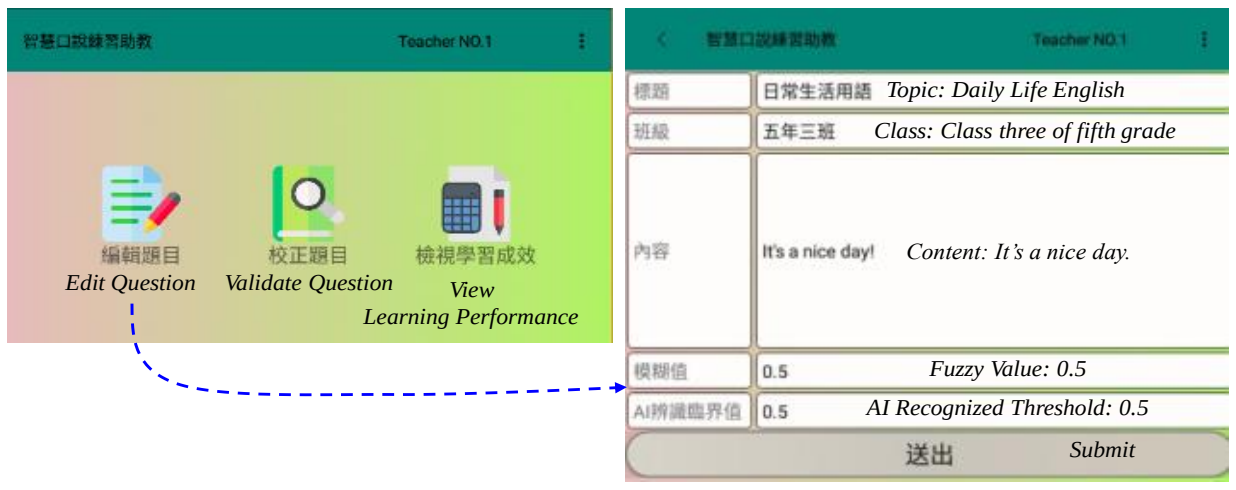

Fig. 7. Screenshots of the home screen of the teacher sub-system and editing question.

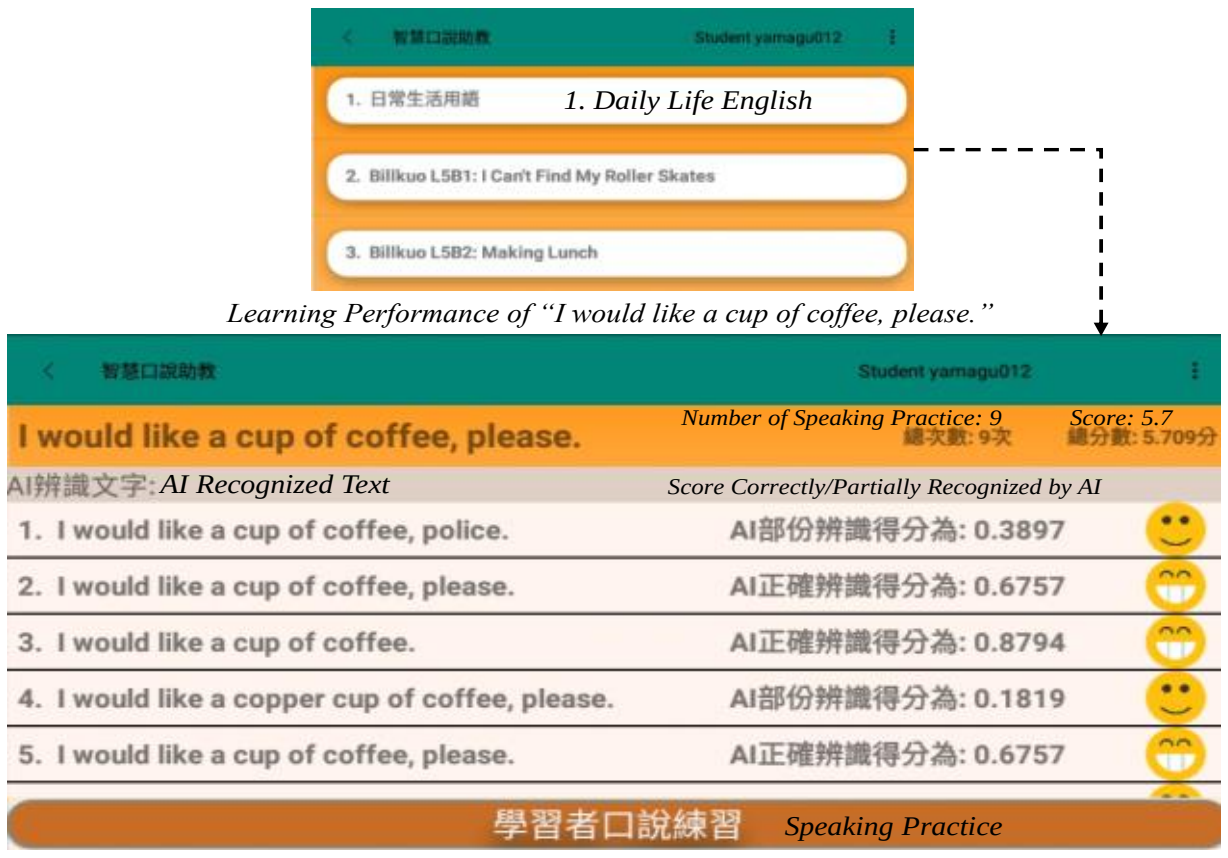

Fig. 8. Screenshots of the choosing/listing learning contents and displaying learning performance of the student sub-system.

Fig. 7 shows the home screen of the developed teacher sub-system, including editing question, validating question, and viewing students' learning performance. And, the screenshot of the editing questions including topic, contents, fuzzy value, and AI recognized threshold. Fig. 8 shows screenshots of choosing/listing the learning contents of the student sub-system. In addition, Fig. 8 shows the screenshot of the learning results, including a correct or partially correct AI recognized score, AI recognized text, and how many times the student speaks this sentence, where happy face and smile face represent that the audio received by the speech recognizer is correctly and partially recognized by AI, respectively.

## IV. Exploration and Exploitation for Student Learning Behavior Ontology Construction

### A. *Student Learning Behavior Ontology*

Fig. 9 shows the observed student learning behavior of AI-FML experimental class in Taiwan from 2019 to 2020. According to [3], the right brain is more humanity; however, the left brain is more rationality. Take an AI-FML experimental course of subject computer science at Rende elementary school in Taiwan for an example. In class, students rationally need to follow human teachers to establish knowledge base, rule base, setup AIoT/Robot environment, …, and execute machine learning tool, in order to interact with the robot Kebbi Air. However, some students (fall-behind or ahead-of-schedule) naturally have a desire for playing game, watching video, …, or listening to music in class. Moreover, some students may feel sleepy sometimes. Fig. 10 shows the ontology structure of the student learning behavior according to the observed student learning behavior in the AI-FML experimental class.

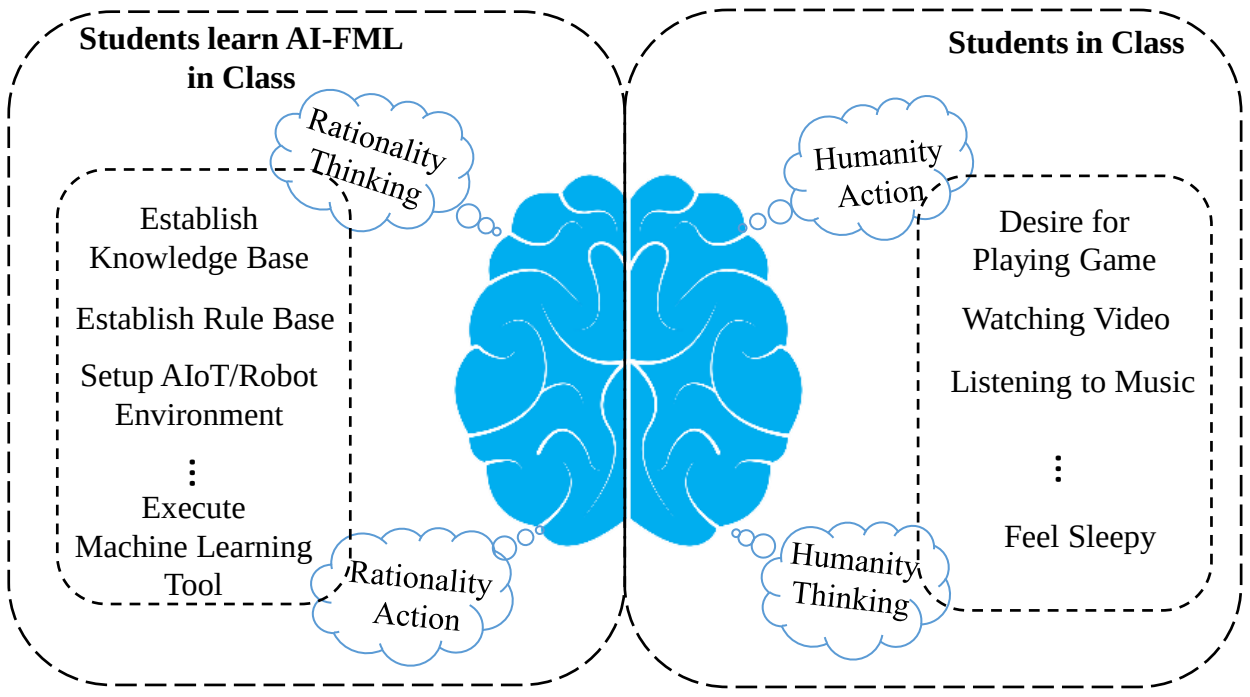

Fig. 9. Student learning behavior of AI-FML experimental class in Taiwan.

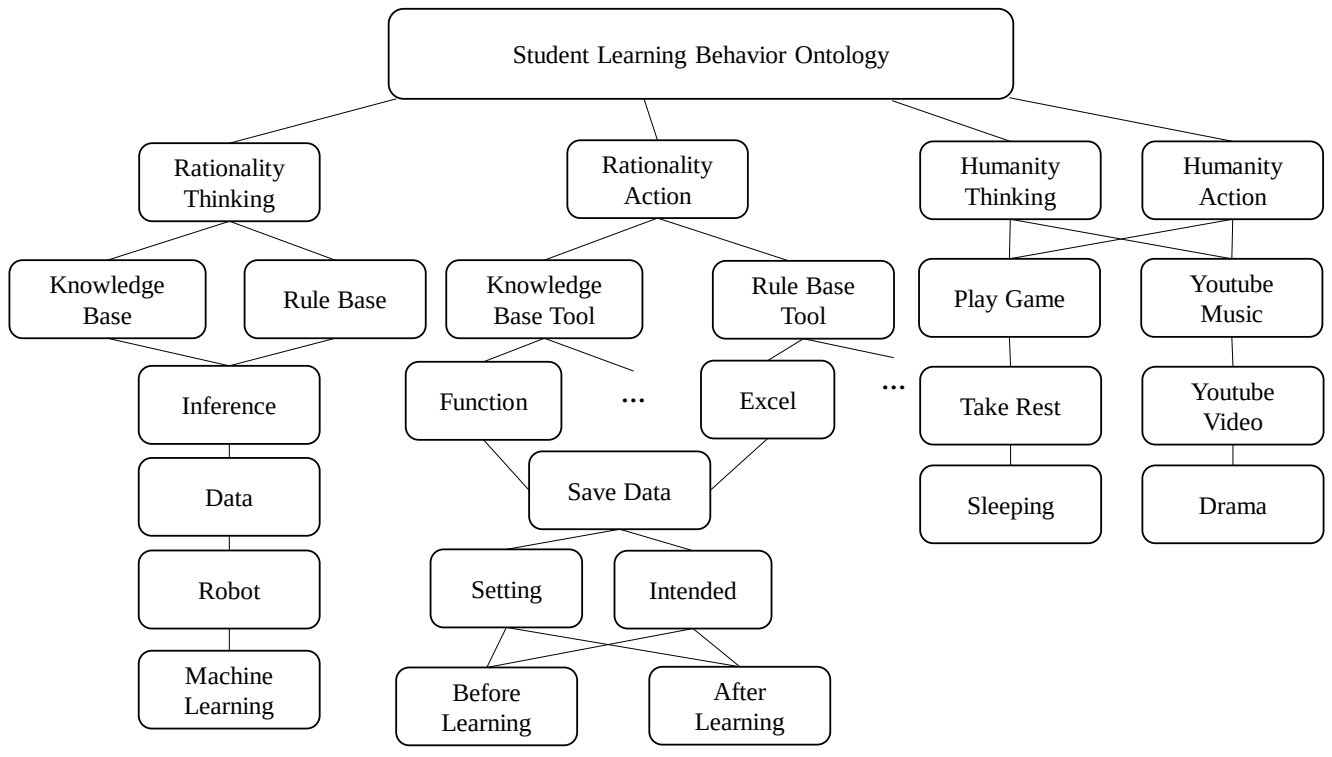

Fig. 10. Student learning behavior ontology for AI-FML experimental class.

### B. *Explore on-line Student Learning Data and Informtion*

In this paper, we adopt the exploration and exploitation strategy to construct student learning behavior ontology for the AI-FML robotic agent [5]. In Taiwan, the introduced AI-FML experimental class time at the elementary school is 40 minutes, and we introduced AI-FML human and machine co-learning class to the Rende elementary school in Tainan, Taiwan in 2019 and 2020. The class is divided into two parts described as follows: Part I (20 minutes) is for human teaching time and Part II (20 minutes) is for AI-FML robotic agent time. In the second part, the AI-FML robotic agent uses its sensors to explore the on-line learning data and information from students in the class. Fig. 11 shows the structure of on-line exploring student learning data and information in class.

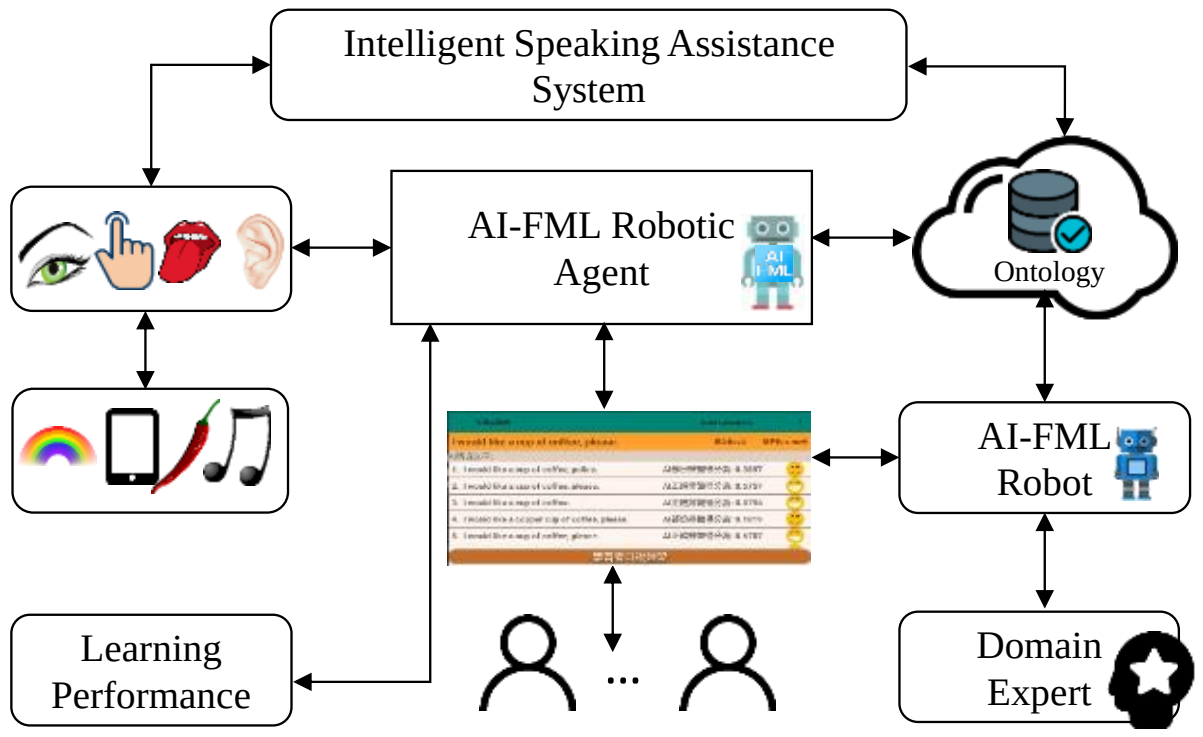

Fig. 11. Explore on-line student learning data and information in class.

### C. *Exploit Student Learned Information and Knowledge to Teach Adaanced Contents*

After students learn English in the classroom via the perception agent, the computational agent in AI-FML robotic agent assess their learning performance based on item response theory (IRT) [6] to justify the learned information and knowledge for each student. Then, the cognition agent sends the learning results to the students or their teachers for learning or teaching advanced content next time.

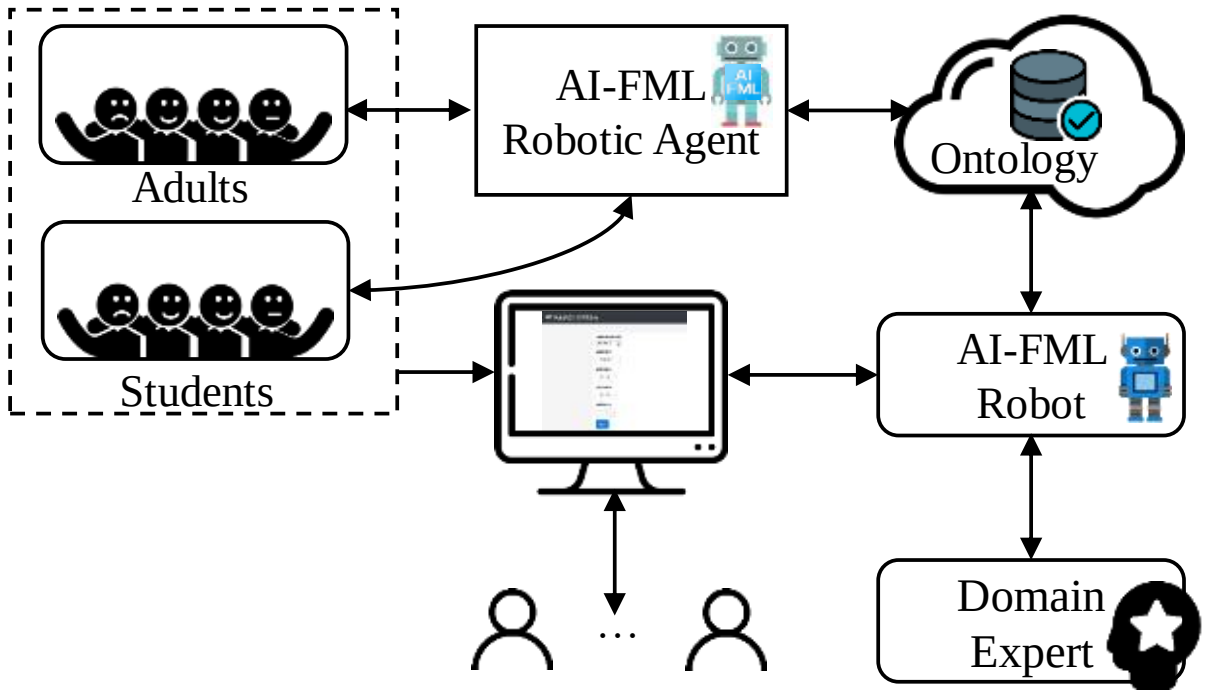

Fig. 12. Exploit student learned information and knowledge for adaptive teaching.

## V. Experimental Results

### A. *Speaking Practice Data in Taiwan*

This subsection describes the speaking practice data in Taiwan in 2019. Based on the system structure of the perception agent and cognition agent shown in Fig. 3, we developed an English speaking practice platform. Since Sept. 2019, National University of Tainan (NUTN) has cooperated with Tainan City Government to introduce AI-FML experimental class of students and robots co-learning English/Computer Science to two elementary schools (ES1: Rende Elementary School and ES2: Guiren Elementary School) and one junior high school (JHS1: Rende Junior High School) in Tainan. Table I shows the partial collected data of JHS1 related to English speaking, where NULL denotes that this student was absent from the class on Oct. 19, 2020. The total number of data from ES1, ES2, and JHS1 is 1110, 2490, and 624. The AI-FML robotic agent summarizes a record with nine feature values, $x_1$, $x_2$, …, and $x_9$, and a label value, $y$, for each student after class. The $x_1$ is school code, for example, JHS1 is 0, ES1 is 1, and ES2 is 2. The $x_2$ is the grade of the involved students. In Taiwan, there are six grades (Grade 1 to Grade 6) at elementary school and three grades (Grade 7 to Grade 9) at junior high school. The $x_3$ is the gender of the involved student. The $x_4$, $x_5$, and $x_6$ are computed scores extracted from his/her feedback. The $x_7$ denotes his/her number of speaking practice in class. The $x_8$ is the ratio correctly recognized by AI. The $x_9$ is the score correctly and partially recognized by AI. The $y_1$ and $y_2$ are his/her monthly English paper-and-pencil test score and listening test score at school, respectively, and the $y$ denotes his/her monthly final exam score which equals $y_1 \times 0.7 + y_2 \times 0.3$. In addition to students of JHS1, the other involved students only have $y_1$ score and their label value $y$ is $y_1$. In this paper, the data pre-processing mechanism, including outlier and standardization analysis, is performed as follows. (1) Replace NULL with the mean value of the respectively collected data. (2) Remove outliers based on measures of the interquartile range (IQR = Q3−Q1) where Q1 and Q3 denote the lower quartile and upper quartile, respectively. When a point is beyond 1.9×IQR, it will be removed from the training dataset. (3) Standardize features $x_4$ to $x_9$ between 0 and 1.

TABLE I. PARTIAL DATA OF PRACTICE SPEAKING FROM JHS1.

| Date | $x_4$ | $x_5$ | $x_6$ | $x_7$ | $x_8$ | $x_9$ | $y_1$ | $y_2$ | $y$ |
|---|---|---|---|---|---|---|---|---|---|
| 2019/10/9 | NULL | NULL | NULL | 5 | 0.4 | 2 | 53 | 81 | 61.4 |
| 2019/10/9 | 8.36 | 8.6 | 5.83 | 12 | 0.417 | 5 | 12 | 24 | 15.6 |
| 2019/10/9 | 9.15 | 9.38 | 9.62 | 16 | 0.25 | 4.476 | 72 | 94 | 78.6 |
| 2019/10/9 | 8.34 | 8.52 | 5.82 | 19 | 0.316 | 7.045 | 46 | 88 | 58.6 |
| | | | | | ⋮ | | | | |
| 2019/10/9 | 9.53 | 9.37 | 9.64 | 16 | 0.063 | 1.681 | 78 | 92 | 82.2 |
| 2019/10/9 | 8.62 | 8.28 | 5.7 | 8 | 0 | 0 | 19 | 66 | 33.1 |
| 2019/10/9 | 9.35 | 9.33 | 9.92 | 11 | 0.182 | 2 | 98 | 100 | 98.6 |
| | | | | | ⋮ | | | | |

Next, we describe the conducted four experimental results of speaking practice performance in Taiwan using DNN model as follows: (1) In Exp. 1, there are seven input features $x_1$ and $x_4$ to $x_9$ in the input layer. The number of nodes in hidden layers 1, 2, 3, 4, and 5 are 21, 64, 512, 64, and 10, respectively, followed by a ReLU activation function. (2) In Exp. 2, there are seven input features $x_1$ and $x_4$ to $x_9$ in the input layer. The number of nodes in hidden layers 1, 2, 3, and 4, are 21, 512, 512, and 3, respectively, followed by a ReLU activation function. (3) In Exp. 3, there are nine input features $x_1$ to $x_9$ in the input layer. The number of nodes in hidden layers 1, 2, 3, 4, and 5 are 27, 64, 512, 64, and 10, respectively, followed by a ReLU activation function. (4) In Exp. 4, there are nine input features $x_1$ to $x_9$ in the input layer. The number of nodes in hidden layers 1, 2, 3, and 4 are 27, 512, 512, and 3, respectively, followed by a ReLU activation function. (5) Exps. 3 and 4 standardize the collected data based on the class date, grade, and school of the involved students.

In all experiments, the final layer has 1 node to denote the predicted learning performance of this semester and applies a regression function. Moreover, the dropout is 0.3, mini-batch is 30, and Adam optimization mechanisms are also adopted in this paper. In addition, the parameters for Adam learning optimization are as follows: learning rate is 0.001, β_1 is 0.1, and β_2 is 0.999. The proportion which splits the training dataset into a validation dataset is 0.3 during training. There are 1,771 training data and 759 testing data after data preprocessing. Mean Square Error (MSE) is used as the loss function. Figs. 13(a) and 13(b) show the loss curves for both the training dataset and validation dataset against iterations 100 and 3000, respectively, in Exp. 1, which indicates that learning for 100 epochs has reached convergence. Fig. 14 shows the loss value after learning 100 epochs in Exps. 1, 2, 3, and 4, which indicates that Exp. 4 has a better performance than the other three experiments.

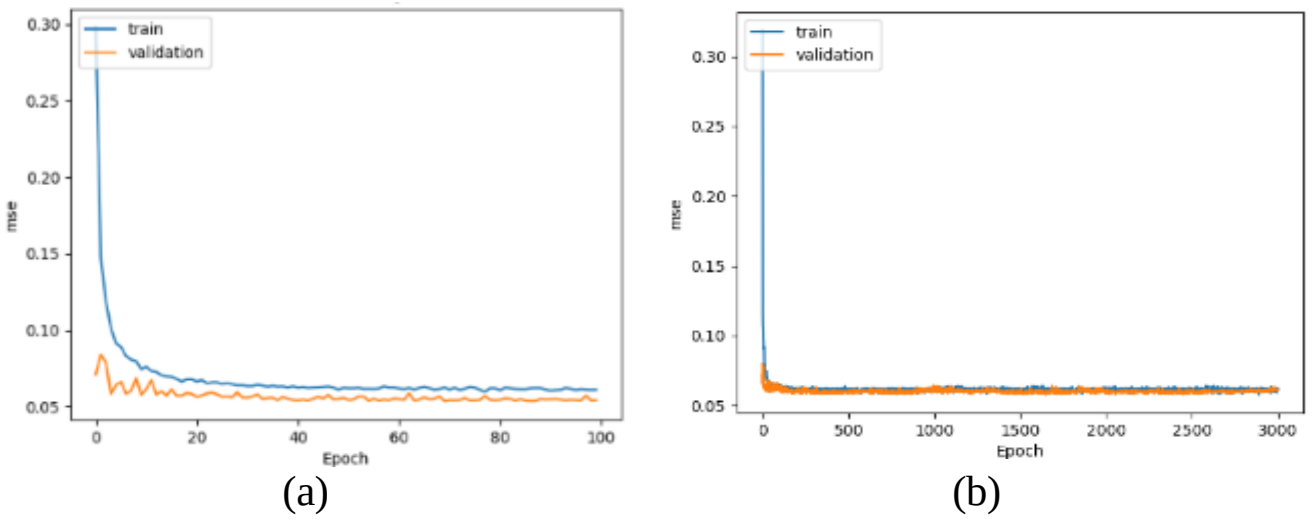

Fig. 13. Loss curve of epochs (a) 100 and (b) 3000.

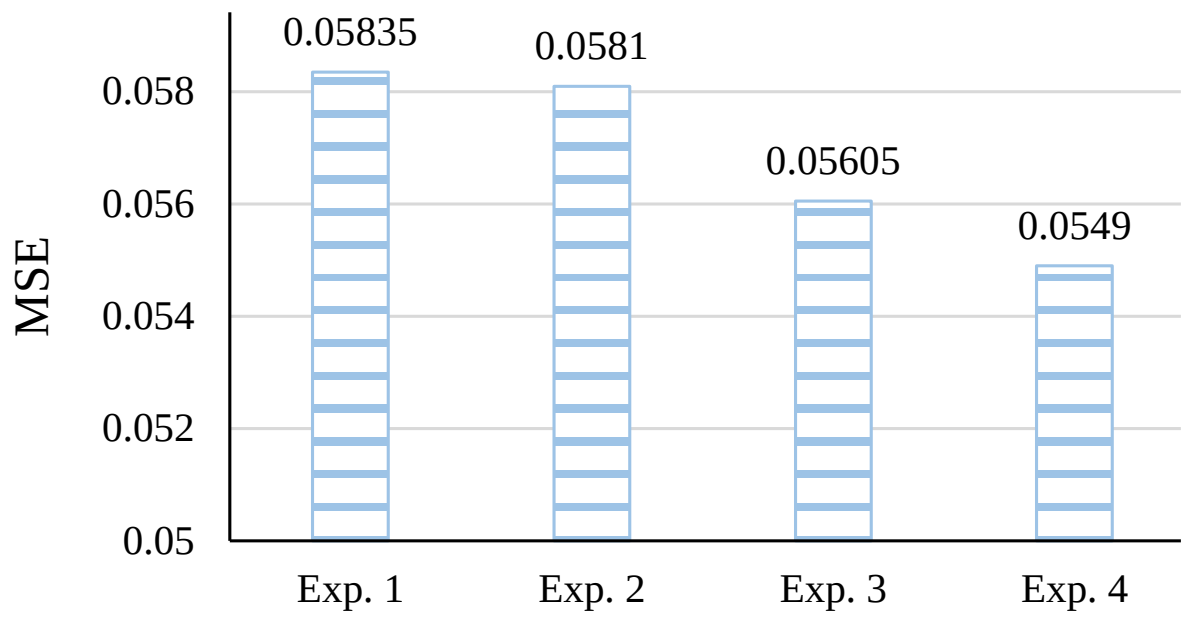

Fig. 14. Loss value after learning 100 epochs in Exps. 1, 2, 3 and 4.

## B. *Speaking Practice Performance in Japan*

NUTN cooperated with Tokyo Metropolitan University (TMU) to collect the learning data of undergraduate and graduate students' English speaking and listening on Feb. 11, 12, and 13, 2020. Table II shows the number of collected data and experimental descriptions on Feb. 11 to Feb. 13 in Japan. Fig. 15 shows the average score recognized by AI in Exp. 5. Sentences "*As soon as possible* and *I would like a cup of coffee, please*" are more difficult for students to speak fluently enough to make AI correctly to be recognized. Fig. 16 shows the average score recognized by AI and ratio correctly recognized by AI of each involved member of Kubota Lab. and Yamaguchi Lab. in Exp. 6. The correctly recognized ratio from AI is about 0.5 for most of the members and yamguchi001 member got the highest average score. In Exp. 7, all members were invited to make an adaptive listening challenge on BillKuo English Levels 1, 2, and 7. The higher the level, the more difficult. Fig. 17 shows that Level 7 is more difficult for most of the involved members than the other two levels but two members, kubota003 and yamaguchi003, did not make a challenge on Level 7 in Exp. 7.

TABLE II. DESCRIPTIONS OF EXPS. ON FEB. 11 TO FEB. 13 IN JAPAN.

| Date / Exp. No. | Feb. 11 2020 | Feb. 12 2020 | Feb. 13 2020 | Total |
|---|---|---|---|---|
| Exp. 5: Speaking Practice with Robot Kebbi Air | 10 | 90 | 70 | 170 |
| Exp. 6: Speaking Practice using an Android Tablet | 30 | 270 | 210 | 510 |
| Exp. 7: Adaptive Challenge using a Notebook | 20 | 180 | 140 | 340 |

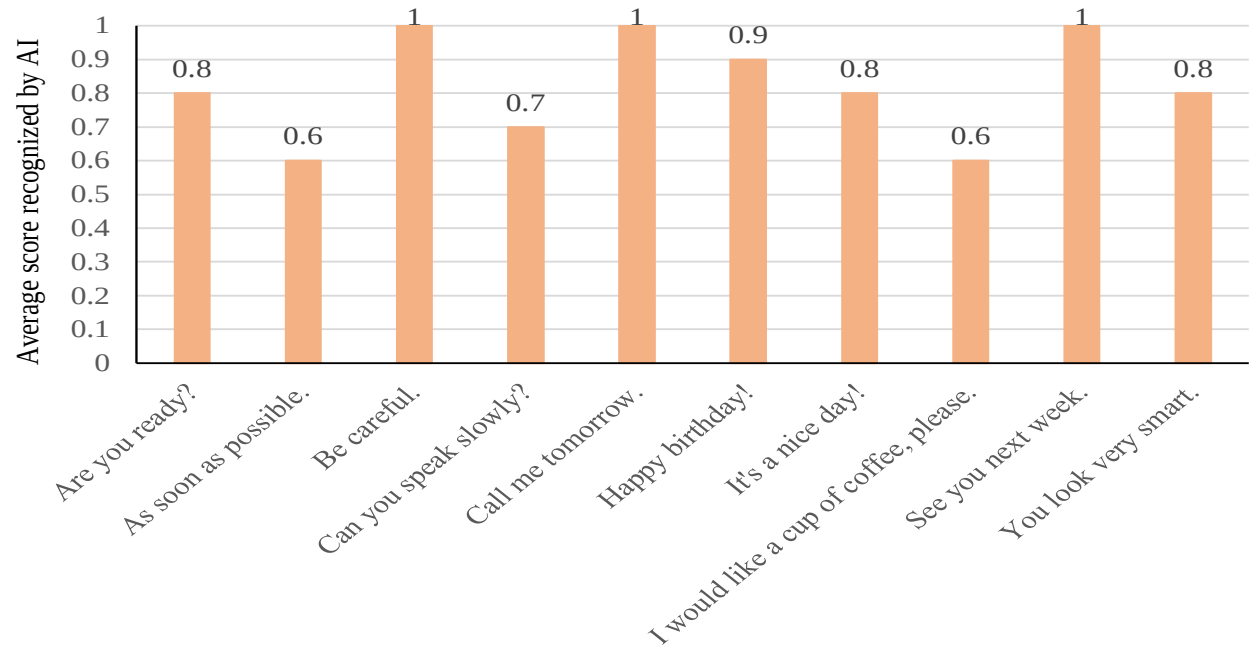

Fig. 15. Average score recognized by AI in Exp. 5.

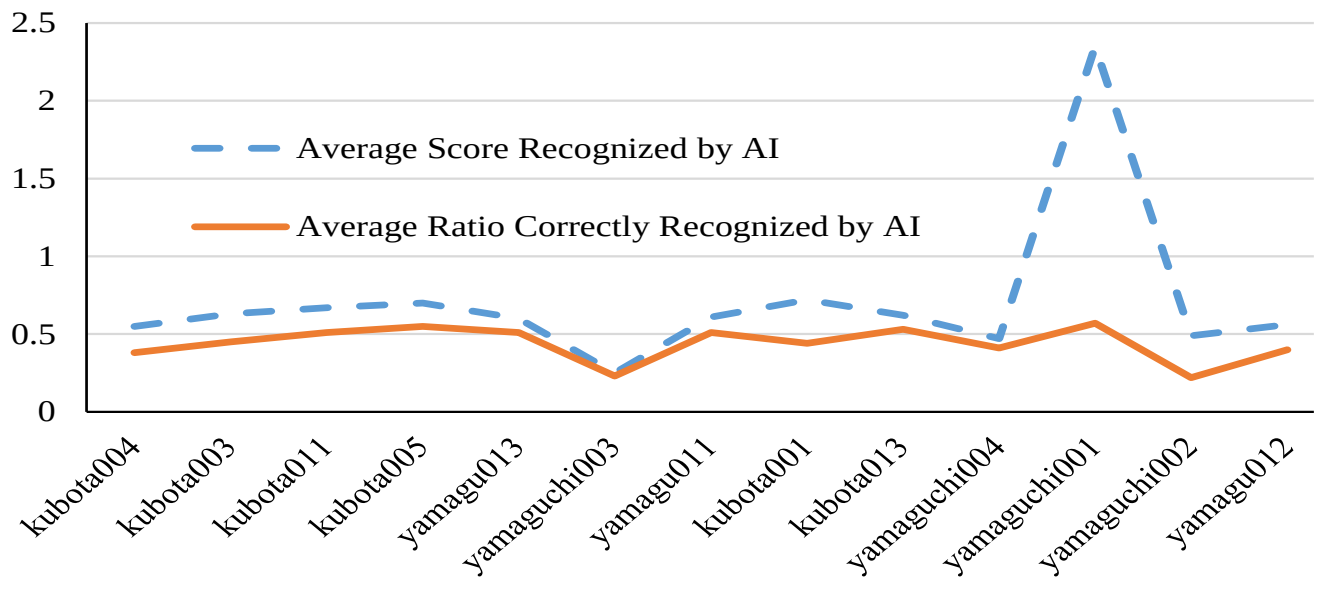

Fig. 16. Average score recognized by AI and ratio correctly recognized by AI in Exp. 6.

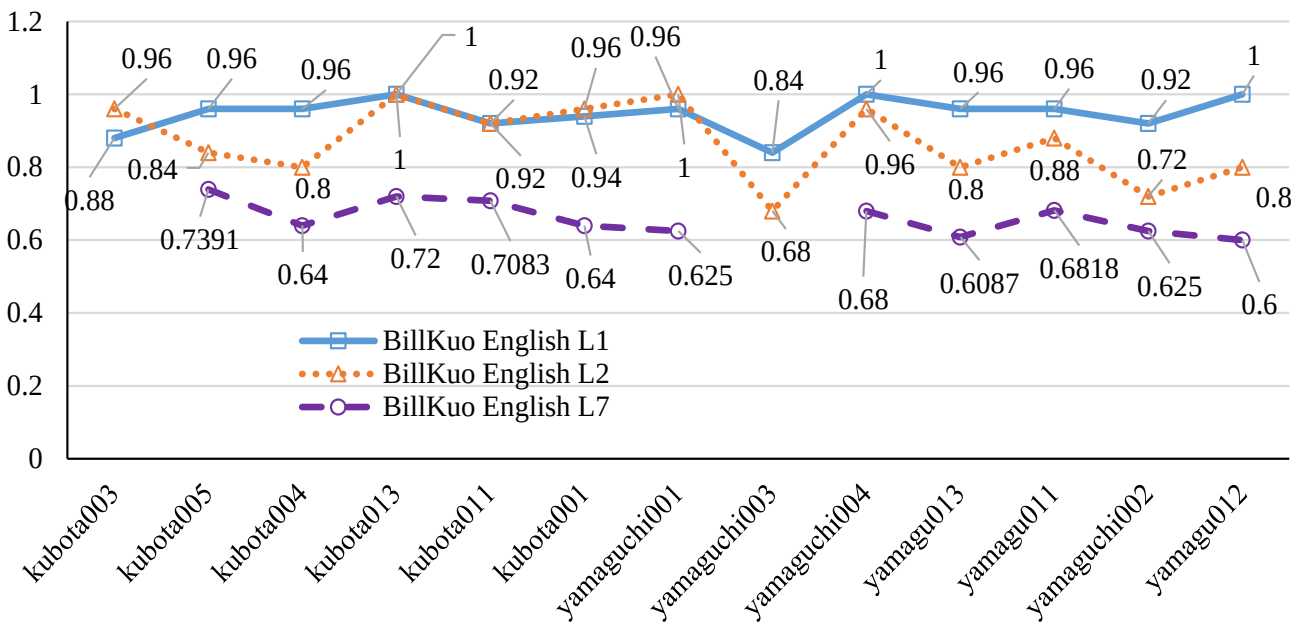

Fig. 17. Ratio correctly answered by involved members when challenging Levels 1, 2, and 7 in Exp. 7.

## VI. CONCLUSIONS

This paper proposes an AI-FML robotic agent for student learning behavior ontology construction, including the perception agent, cognition agent, and computational agent, for analyzing student learning behavior. Moreover, the perception agent and the cognition agent are deployed in the Kebbi Air robot. In addition, the computational agent with the DNN model can communicate with the perception agent and cognition agent via the Internet. The proposed AI-FML robotic agent has been applied to the teaching fields in Taiwan and tested in Japan. The involved students interacted with the Kebbi Air to co-learn English and AI-FML in class which makes learning fun for students. The experimental results show that the agents can be utilized in the human and machine co-learning model for the future education.

## ACKNOWLEDGMENT

The authors would like to thank the staff of Center for Research of Knowledge Application & Web Service (KWS Center) of NUTN and the involved faculty and students of Rende elementary school, Guiren elementary school, and Rende junior high school in Taiwan, and the members of Yamaguchi Lab. and Kubota Lab. of TMU in Japan.

## REFERENCES

[1] IEEE CIS, "What is CI?," May 2020. [Online] Available: https://cis.ieee.org/about/what-is-ci.

[2] B. Bouchon-Meunier, "Welcome from the President of IEEE CIS," May 2020. [Online] Available: https://cis.ieee.org/about.

[3] Healthline, "Left brain vs. right brain: what does this mean for me?" May 2020. [Online] Available: https://www.healthline.com/health/left-brain-vs-right-brain.

[4] P. Soulos, "Agent perception," May 2020. [Oneline] Available: http://paulsoulos.com/editorial/2016/08/10/agent-perception-and-learning.html.

[5] B. Christian and T. Griffiths, "Algorithms to Live By: The Computer Science of Human Decisions," Picador, USA, 2017.

[6] C. S. Lee, M. H. Wang, C. S. Wang, O. Teytaud, J. L. Liu, S. W. Lin, and P. H. Hung, "PSO-based fuzzy markup language for student learning performance evaluation and educational application," *IEEE Transactions on Fuzzy Systems*, vol. 26, no. 5, pp. 2618-2633, 2018.

[7] C. S. Lee, M. H. Wang, Y. L. Tsai, L. W. Ko, B. Y. Tsai, P. H. Hung, L. A. Lin, and N. Kubota, "Intelligent agent for real-world applications on robotic edutainment and humanized co-learning," *Journal of Ambient Intelligence and Humanized Computing*, 2019. (DOI: 10.1007/s12652-019-01454-4)

[8] C. S. Lee, Y. L. Tsai, M. H. Wang, W. K. Kuan, Z. H. Ciou, and N. Kubota, "AI-FML agent for robotic game of Go and AIoT real-world co-learning applications," *2020 World Congress on Computational Intelligence* (*IEEE WCCI 2020*), Glasgow, Scotland, UK, Jul. 19-24, 2020. (Accepted)

[9] C. S. Lee, Y. L. Tsai, M. H. Wang, H. Sekino, T. X. Huang, W. F. Hsieh, E. Sato-Shimokawara, and T. Yamaguchi, "FML-based machine learning tool for human emotional agent with BCI on music application," *2019 International Conference on Technologies and Applications of Artificial Intelligence* (*TAAI 2019*), Kaohsiung, Taiwan, Nov. 21-23, 2019.

[10] C. S. Lee, M. H. Wang, Y. L. Tsai, R. P. Chang, L. C. Li, N. Takese, S. Yamamoto, and N. Kubota, "FML-based intelligent agent for robotic e-learning and entertainment application," *2019 International Conference on Technologies and Applications of Artificial Intelligence* (*TAAI 2019*), Kaohsiung, Taiwan, Nov. 21-23